\documentclass[a4paper,10pt,twoside]{article}

\usepackage{clin}        
\usepackage{harvard}     
\usepackage{amsmath} 
\usepackage{amssymb}  
\usepackage{graphics}
\usepackage[graphicx]{realboxes}
\usepackage{subcaption}

\begin{document}

\title{Binary and Multitask Classification Model for Dutch Anaphora Resolution: Die/Dat Prediction}

\author{Liesbeth Allein$^*$ \email{liesbeth.allein@kuleuven.be}\\
{\normalsize \bf Artuur Leeuwenberg}$^{**}$ \email{A.M.Leeuwenberg-15@umcutrecht.nl}\\
{\normalsize \bf Marie-Francine Moens}$^{*}$ \email{sien.moens@kuleuven.be}
\AND \addr{$^*$Department of Computer Science, KU Leuven, Celestijnenlaan 200A, Leuven, Belgium}
\AND \addr{$^{**}$Julius Center, University Medical Center Utrecht, Utrecht, The Netherlands} }

\maketitle\thispagestyle{empty} 


\begin{abstract}
The correct use of Dutch pronouns \textit{die} and \textit{dat} is a stumbling block for both native and non-native speakers of Dutch due to the multiplicity of syntactic functions and the dependency on the antecedent's gender and number. Drawing on previous research conducted on neural context-dependent dt-mistake correction models \cite{heyman2018automatic}, this study constructs the first neural network model for Dutch demonstrative and relative pronoun resolution that specifically focuses on the correction and part-of-speech prediction of those two pronouns. Two separate datasets are built with sentences obtained from, respectively, the Dutch Europarl corpus \cite{koehn2005europarl} - which contains the proceedings of the European Parliament from 1996 to the present - and the SoNaR corpus \cite{oostdijk2013construction} - which contains Dutch texts from a variety of domains such as newspapers, blogs and legal texts. Firstly, a binary classification model solely predicts the correct \textit{die} or \textit{dat}. The classifier with a bidirectional long short-term memory architecture achieves 84.56\% accuracy. Secondly, a multitask classification model simultaneously predicts the correct \textit{die} or \textit{dat} and its part-of-speech tag. The model containing a combination of a sentence and context encoder with both a bidirectional long short-term memory architecture results in 88.63\% accuracy for \textit{die/dat} prediction and 87.73\% accuracy for part-of-speech prediction. More evenly-balanced data, larger word embeddings, an extra bidirectional long short-term memory layer and integrated part-of-speech knowledge positively affects \textit{die/dat} prediction performance, while a context encoder architecture raises part-of-speech prediction performance. This study shows promising results and can serve as a starting point for future research on machine learning models for Dutch anaphora resolution.
\end{abstract}

\section{Introduction}

Following previous research on automatic detection and correction of dt-mistakes in Dutch \cite{heyman2018automatic}, this paper investigates another stumbling block for both native and non-native speakers of Dutch: the correct use of \textit{die} and \textit{dat}. The multiplicity of syntactic functions and the dependency on the antecedent's gender and number make this a challenging task for both human and computer. The grammar concerning \textit{die} and \textit{dat} is threefold. Firstly, they can be used as dependent or independent demonstrative pronouns \textit{(aanwijzend voornaamwoord)}, with the first replacing the article before the noun it modifies and the latter being a noun phrase that refers to a preceding/following noun phrase or sentence. 
The choice between \textit{die} and \textit{dat} depends on the gender and number of the antecedent: \textit{dat} refers to neuter, singular nouns and sentences, while \textit{die} refers to masculine, singular nouns and plural nouns independent of their gender. Secondly, \textit{die} and \textit{dat} can be used as relative pronouns introducing relative clauses \textit{(betrekkelijk voornaamwoord)}, which provide additional information about the directly preceding antecedent it modifies. 
Similar rules as for demonstrative pronouns apply: masculine, singular nouns and plural nouns are followed by relative pronoun \textit{die}, neuter singular nouns by \textit{dat}. Lastly, \textit{dat} can be used as a subordinating conjunction \textit{(onderschikkend voegwoord)} introducing a subordinating clause. A brief overview of the grammar is given in Table \ref{tab:grammar}. 

The aim is to develop (1) a binary classification model that automatically detects, predicts and corrects \textit{die} and \textit{dat} instances in texts and (2) a multitask classification model that jointly predicts the correct \textit{die/dat} instance and its syntactic function. Whereas research on neural-based, machine learning approaches for Dutch demonstrative and relative pronoun resolution - especially for \textit{die} and \textit{dat} - is to our knowledge non-existing, this paper is a starting point for further research on machine learning applications concerning Dutch subordinating conjunctions, demonstrative pronouns and relative pronouns.

\begin{table}[]
    \centering
    \begin{tabular}{|c|c|c|c|}
        \hline
        Function & Demonstrative & Relative & Subordinating \\
          & pronoun & pronoun & conjunction \\
        \hline
        Refer to antecedent & & & \\
        \textit{singular, masculine noun} & die & die & - \\
        \textit{singular, neuter noun} & dat & dat & - \\
        \textit{plural noun} & die & die & - \\
        \textit{sentence} & dat & - & - \\
        \hline
        Introduce subordinating clause & - & - & dat \\
        \hline
    \end{tabular}
    \caption{Grammar concerning \textit{die} and \textit{dat}}
    \label{tab:grammar}
\end{table}

\section{Related Work} \label{Related Work}

The incentive for this paper is the detection and correction system for dt-mistakes in Dutch \cite{heyman2018automatic}. For that task, a system with a context encoder - a bidirectional LSTM with attention mechanism - and verb encoder - of which the outputs are then fed to a feedforward neural network - has been developed to predict different verb suffixes. As mentioned above, this paper explores the possibility of constructing a neural network system for correcting Dutch demonstrative and relative pronouns \textit{die} and \textit{dat}. The task is also called pronoun resolution or anaphora resolution. Anaphora resolution and pronoun prediction has been major research subjects in machine translation research. Novák et al. (2015), for example, studied the effect of multiple English coreference resolvers on the pronoun translation in English-Dutch machine translation system with deep transfer has been investigated. Niton, Morawiecki and Ogrodnizuk (2018) developed a fully connected network with three layers in combination with a sieve-based architecture for Polish coreference resolution \cite{niton2018deep}. Not only in machine translation, but also in general natural language processing much research has been conducted on machine learning approaches towards coreference resolution \cite{ng2002improving,culotta2007first,zhekova2010ubiu} and pronoun resolution \cite{strube2003machine,zhao2007identification}. However, little to no research has been conducted specifically on \textit{die/dat} correction.

\section{Datasets}

The datasets used for training, validation and testing contain sentences extracted from the Europarl corpus \cite{koehn2005europarl} and SoNaR corpus \cite{oostdijk2013construction}. The Europarl corpus is an open-source parallel corpus containing proceedings of the European Parliament. The Dutch section consists of 2,333,816 sentences and 53,487,257 words. The SoNaR corpus comprises two corpora: SONAR500 and SONAR1. The SONAR500 corpus consists of more than 500 million words obtained from different domains. Examples of text types are newsletters, newspaper articles, legal texts, subtitles and blog posts. All texts except texts from social media have been automatically tokenized, POS tagged and lemmatized. It contains significantly more data and more varied data than the Europarl corpus. Due to the high amount of data in the corpus, only three subparts are used: Wikipedia texts, reports and newspaper articles. These subparts are chosen because the number of wrongly used \textit{die} and \textit{dat} is expected to be low.

\section{Preprocessing} \label{preprocessing}

The sentences in the Europarl corpus are tokenized and parsed using the Dutch version of TreeTagger \cite{schmid1994probabilistic}. Only sentences which contain at least one \textit{die} or \textit{dat} are extracted from the corpora. Subsequently, each single occurrence of \textit{die} and \textit{dat} is detected and replaced by a unique token ('PREDICT'). When there are multiple occurrences in one sentence, only one occurrence is replaced at a time. Consequently, a sentence can appear multiple times in the training and test dataset with the unique token for \textit{die} and \textit{dat} at a different place in the sentence. Each sentence is paired with its automatically assigned ground truth label for \textit{die} and \textit{dat}. The resulting datasets consist of 103,871 (Europarl) and 1,269,091 (SoNaR) sentences. The Europarl dataset, on the one hand, contains 70,057 \textit{dat}-labeled and 33,814 \textit{die}-labeled sentences. The SoNaR dataset, on the other hand, has more than ten times the number of labeled sentences with 736,987 \textit{dat}-labeled and 532,104 \textit{die}-labeled. Considering the imbalance in both datasets, it may be argued that \textit{dat} occurs more frequently than \textit{die} due to its syntactic function as subordinating conjunction and not to its use as demonstrative pronoun whereas it can only refer to singular, neuter nouns. As for the multitask classification model, the POS tags for \textit{die} and \textit{dat} present in the SoNaR corpus are extracted and stored as ground truth labels: 407,848 \textit{subordinating conjunction}, 387,292 \textit{relative pronoun} and 473,951 \textit{demonstrative pronoun}. From a brief qualitative assessment on the POS tags for \textit{die} and \textit{dat} in both corpora, the POS tags in the SoNaR corpus appear to be more reliable than the POS tags generated by TreeTagger in the Europarl corpus. Therefore, only the SoNaR dataset is used for the multitask classification. An overview of the datasets after preprocessing is given in Table \ref{tab:dataset}.

\begin{table}[]
    \centering
    \begin{tabular}{|c|c|c|c|}
        \hline
        Dataset & \# sentences & \textit{dat/} & \textit{subordinating conjunction/} \\
         & & \textit{die} & \textit{relative pronoun/} \\
         & & & \textit{demonstrative pronoun} \\
        \hline
        Europarl & 103,871 & 70,057/ & - \\
         & & 33,814 & - \\
        \hline
        SoNaR & 1,269,091 & 736,987/ & 407,848/ \\
         & & 532,104 & 387,292/ \\
         & & & 473,951 \\
        \hline
    \end{tabular}
    \caption{Overview of datasets}
    \label{tab:dataset}
\end{table}

\section{Binary Classification Model} \label{binary_classification_model}

\subsection{Model Architecture}

\begin{figure}
    \centering
    \includegraphics[width = 7 cm]{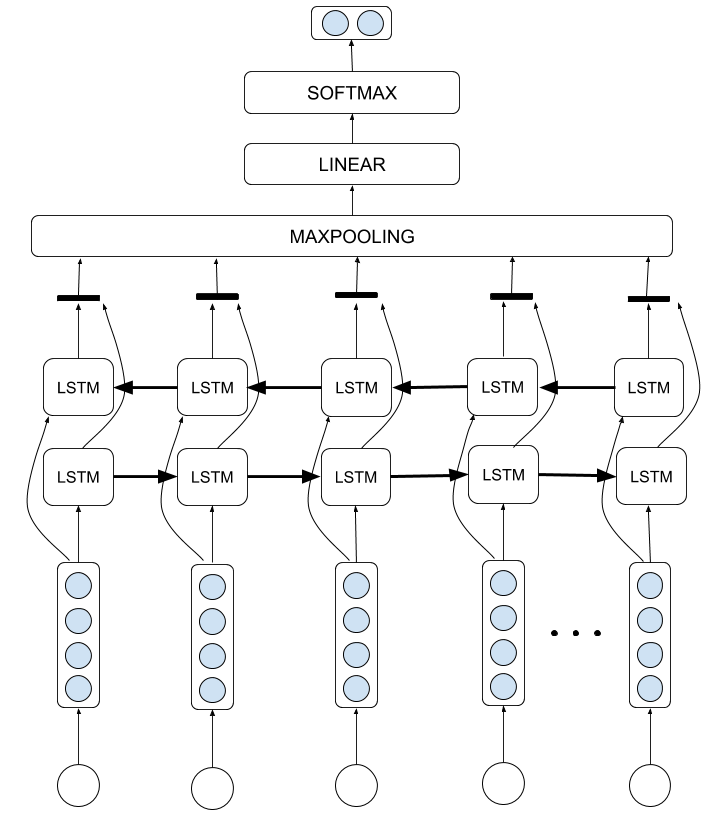}
    \caption{Model architecture of the binary classification model}
    \label{fig:binary_classification}
\end{figure}

For the binary classification model that predicts the correct \textit{die} or \textit{dat} for each sentence, a Bidirectional Long-Short Term Memory (BiLSTM) neural network is deployed. Whereas the antecedent can be rather distant from the demonstrative pronoun due to adjectives and sentence boundaries, an LSTM architecture is chosen over a regular Recurrent Neural Network as the latter does not cope well with learning non-trivial long-distance dependencies \cite{chiu2016named}. Furthermore, a bidirectional LSTM is chosen over a single left-to-right LSTM, whereas the antecedent can be either before or after the \textit{die} or \textit{dat}. The architecture of the binary classification model is provided in Fig. \ref{fig:binary_classification}. The input sentence is first sent through an embedding layer where each token is transformed to a 100-dimensional word embedding which has been initially trained on the dataset of sentences containing at least one \textit{die} or \textit{dat} using the Word2Vec Skip-gram model \cite{mikolov2013distributed}. The weights of the embedding layer are trainable. The word embeddings are then sent through a BiLSTM layer. The BiLSTM concatenates the outputs of two LSTMs: the left-to-right $LSTM_{forward}$ computes the states $\overrightarrow{h_1}..\overrightarrow{h_N}$ and the right-to-left $LSTM_{backward}$ computes the states $\overleftarrow{h_N}..\overleftarrow{h_1}$. This means that at time $t$ for input $x$, represented by its word embedding $E(x)$, the bidirectional LSTM outputs the following:

\begin{equation}
    h_t = [ \overrightarrow{h_t} ;  \overleftarrow{h_t} ] \footnote{ [ ; ] denotes concatenation}
\end{equation}
\begin{equation}
    \overrightarrow{h_t} = LSTM_{forward}(\overrightarrow{h_{t-1}},E(x_t))
\end{equation}
\begin{equation}
    \overleftarrow{h_t} = LSTM_{backward}(\overleftarrow{h_{t+1}},E(x_t))
\end{equation}

Next, the concatenated output is sent through a maxpooling layer, linear layer and, eventually, a softmax layer that generates a probability distribution over the two classes. In order to prevent the model from overfitting and co-adapting too much, dropout regularization is implemented in the embedding layer and the linear layer. In both layers, dropout is set to \(p = 0.5\) which randomly zeroes out nodes in the layer using samples from a Bernoulli distribution.

\subsection{Experimental Set-Up}

Each dataset is randomly divided into a training (70\%), validation (15\%) and test set (15\%). The data is fed to the model in batches of 128 samples and reshuffled at every epoch. The objective function that is minimized is Binary Cross-Entropy:

\begin{equation}
    BCE_{p}(q) = -\frac{1}{N}\sum_{i=1}^{N}y_i \cdot log(p(\hat{y}_i)) + (1 - y_i) \cdot log(1 - p(\hat{y}_i))
\end{equation}

where \(y_i\) is the ground truth label (0 for \textit{dat} and 1 for \textit{die}) and \(p(\hat{y}_i)\) is the probability of the predicted label for all \(N\) input sentences of the train set. The weights are optimized using Stochastic Gradient Descent with learning rate = 0.01 and momentum = 0.9. The data is fed to the model in 24 epochs. 

\subsection{Results}

An overview of the performance results is given in Table \ref{tab:tests_binary}. We compare model performance when trained and tested on the two corpora individually and experiment with different settings of the two corpora in order to investigate the effect of dataset changes on model performance. There are three settings: \textit{full} in which the datasets contain full sentences, \textit{windowed} in which sentences are windowed around the unique prediction token without exceeding sentence boundaries (max. five tokens before and after the token, including token), and \textit{windowed no\_boundaries} in which the windows can exceed sentence boundaries. When limiting the input sentences to windowed sentences in the Europarl corpus (2), model performance increases significantly on all metrics, especially for \textit{die} prediction performance. The difference in model performance when trained and tested on the Europarl (2) and SoNaR (3) windowed datasets is particularly noticeable in the precision, recall and F1 scores. Model performance for \textit{dat} prediction is better for the Europarl dataset than for the SoNaR dataset, while model performance for \textit{die} prediction is notably better for the SoNaR dataset than for the Europarl dataset. Lastly, a change in windowing seems to have a positive impact on the overall model performance: the model trained and tested on the SoNaR dataset with windows exceeding sentence boundaries (3) outperforms the model trained and tested on the SoNaR dataset with windows within sentence boundaries (4) on every metric.

\begin{table}[]
    \centering
    \begin{tabular}{|c|c|c|c|c|c|}
        \hline
        \multicolumn{6}{|c|}{Binary Classification Model} \\
         \hline
         Dataset & Accuracy & Balanced & Precision & Recall & F1 \\
          & & accuracy & dat/die & dat/die & dat/die \\
         \hline
         \hline
         Europarl, \textit{full} (1) & 75.03\% & 68.49\% & 78.11\%/ & 87.45\%/ & 82.41\%/ \\
          & & & 65.68\% & 49.54\% & 56.05\% \\ 
         \hline
         Europarl, \textit{windowed} (2) & 83.27\% & 80.70\% & 87.19\%/ & \textbf{88.14\%}/ & \textbf{87.58\%}/ \\
          & & & 74.97\% & 73.26\% & 73.83\%\\
         \hline
         SoNaR, \textit{windowed} (3) & 82.34\% & 81.72\% & 85.35\%/ & 84.94\%/ & 85.06\%/ \\
          & & & 77.94\% & 78.50\% & 78.05\% \\ 
         \hline
         SoNaR, \textit{windowed}, \textit{no\_boundaries} (4) & \textbf{84.56\%} & \textbf{84.18\%} & \textbf{87.71}\%/ & 86.16\%/ & 86.85\%/ \\
          & & & \textbf{80.13\%} & \textbf{82.20\%} & \textbf{80.99\%} \\ 
         \hline
    \end{tabular}
    \caption{Performance results of the binary classification model on the Europarl dataset containing full sentences (1), the Europarl dataset containing windowed sentences within sentence boundaries (2), the SoNaR dataset containing windowed sentences within sentence boundaries (3) and the SoNaR dataset containing windowed sentences exceeding sentence boundaries (4).}
    \label{tab:tests_binary}
\end{table}

\section{Multitask Classification Model} \label{multitask_classification_model}

\subsection{Model Architecture}

\begin{figure}
\begin{subfigure}{.5\textwidth}
    \centering
    \includegraphics[width = 7cm]{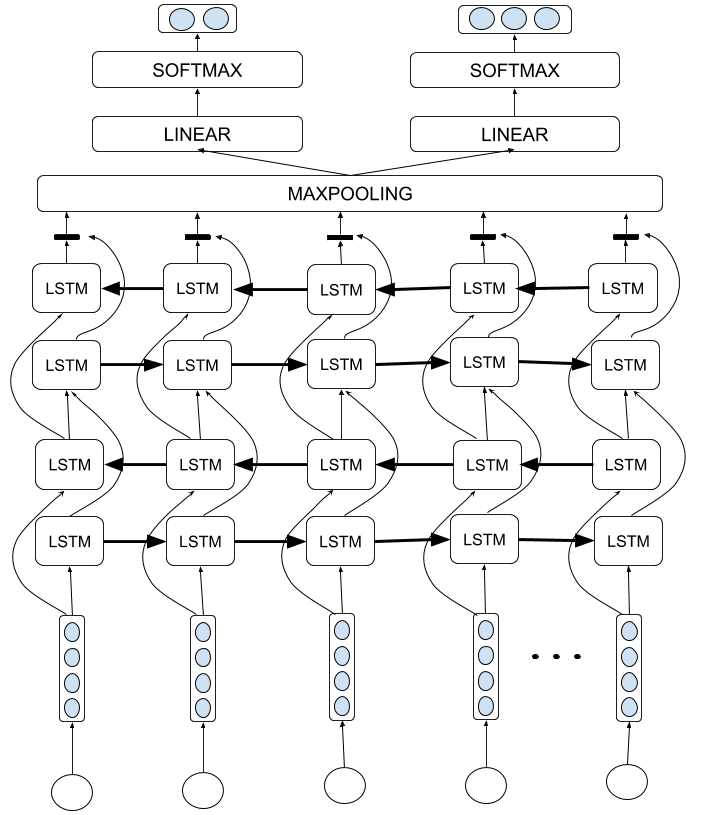}
    \caption{BiLSTM model}
    \label{fig:mlt_classification_1}
\end{subfigure}
\begin{subfigure}{.5\textwidth}
    \centering
    \includegraphics[width = 7.25cm]{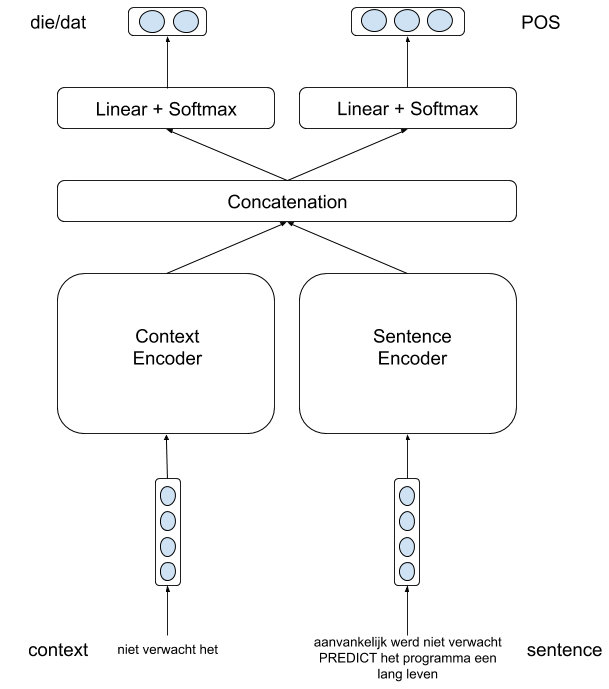}
    \caption{Context + Sentence Encoder}
    \label{fig:mlt_classification_2}
\end{subfigure}
\caption{Overview of the two multitask classification model architectures}
\label{fig:mlt_classification}
\end{figure}

The second model performs two prediction tasks. The first prediction task remains the binary classification of \textit{die} and \textit{dat}. The second prediction task concerns the prediction of three parts-of-speech (POS) or word classes, namely \textit{subordinating conjunction}, \textit{relative pronoun} and \textit{demonstrative pronoun}. An overview of the model architectures is given in Fig. \ref{fig:mlt_classification}. For the BiLSTM model, the first layer is the embedding layer where the weights are initialized by means of the 200-dimensional pre-trained embedding matrix. The weights are updated after every epoch. The second layer consists of two bidirectional LSTMs where the output of the first BiLSTM serves as input to the second BiLSTM. The layer has dropout regularization equal to 0.2. The two-layer BiLSTM layer concatenates the outputs at time \(t\) into a 64-dimensional vector and sends it through a maxpooling layer. Until this point, the two tasks share the same parameters. The model then splits into two separate linear layers. The left linear layer transforms the 64-dimensional vector to a two-dimensional vector on which the softmax is computed. That softmax layer outputs the probability distribution over the \textit{dat} and \textit{die} labels. The right linear layer transforms the 64-dimensional vector to a three-dimensional vector on which a softmax function is applied. The softmax layer outputs the probability distribution over the \textit{subordinating conjunction}, \textit{relative pronoun} and \textit{demonstrative pronoun} labels. The second multitask classification model takes the immediate context around the 'PREDICT' token (two tokens before and one token after) as additional input. Both the windowed sentence and context are first transformed into their word embedding representations. They are then sent through a sentence encoder and context encoder, respectively. The sentence encoder has the same architecture as the second and third layer of the BiLSTM model, namely a two-layer BiLSTM and a maxpooling layer. For the context encoder, we experiment with two different architectures: a feedforward neural network and a one-layer BiLSTM with dropout = 0.2 with a maxpooling layer on top. Both sentence and context encoder output a 64-dimensional vector which are, consequently, concatenated to a 128-dimensional vector. As in the BiLSTM model, the resulting vector is sent through two separate linear layers to output probability distributions for both the \textit{die/dat} and POS prediction task.

\subsection{Experimental Set-up}

As discussed in Section \ref{preprocessing}, the POS ground truth labels in SoNaR-based datasets are more reliable than the POS labels in the Europarl-based datasets that are generated by TreeTagger. Consequently, only the SoNaR dataset is used for training and testing. The dataset is randomly divided into a training (70\%), validation (15\%) and test (15\%) set. The data is fed into the model in batches of 516 samples and the data is reshuffled at every epoch. For \textit{die/dat} prediction, the Binary Cross-Entropy loss function is minimized. The weights are optimized using Stochastic Gradient Descent with learning rate = 0.01 and momentum = 0.9. For POS prediction, Cross-Entropy is minimized:
\begin{equation}
    CE(\theta) = -\sum^{C}_{c=1} y_{i,c}log(p_{i,c})  
\end{equation}
where \(C\) is the number of classes (in this case three) \(y_{i,c}\) is the binary indicator (1 or 0) if class label \(c\) is the correct predicted classification for input sentence \(i\) or not, and \(p\) is the probability of sentence \(i\) having class label \(c\). The weights are optimized using Adam optimization with learning rate being equal to 0.0001. The data is fed to the model in 35 epochs.

\subsection{Results}

An overview of the performance results for \textit{die/dat} prediction is given in Table \ref{tab:mlt_diedat}. The same dataset settings as for the binary classification model are used: \textit{full} in which the datasets contain full sentences, \textit{windowed} in which sentences are windowed around the unique prediction token without exceeding sentence boundaries (max. five tokens before and after the token, including token), and \textit{windowed no\_boundaries} in which the windows can exceed sentence boundaries. As mentioned in section \ref{preprocessing}, we only use the SoNaR dataset. The multitask classification models generally perform better with the \textit{windowed} and \textit{windowed no\_boundaries} dataset setting for \textit{die/dat} prediction. Concerning the model architectures, it can be concluded that altering the model architecture has no large impact on model performance for \textit{die/dat} prediction. However, altering the model architecture from an architecture with merely a sentence encoder to an architecture with both a sentence and a context encoder does have a more significant positive impact on model performance for POS prediction (Table \ref{tab:mlt_functions}). For that prediction task, the multitask classification model with a BiLSTM context encoder trained and tested on \textit{windowed} SoNaR sentences reaches best performance results on almost all evaluation metrics. 

\begin{table}[]
    \centering
    \begin{tabular}{|c|c|c|c|c|c|}
        \hline
         Dataset & Accuracy & Balanced & Precision & Recall & F1 \\
          & & accuracy & dat/die & dat/die & dat/die \\
         \hline
         \multicolumn{6}{|c|}{Multitask Classification Model: BiLSTM (1)} \\
         \hline
         SoNaR, \textit{full} & 78.52\% & 77.56\% & 81.59\%/ & 82.60\%/ & 82.06\%/ \\
          & & & 73.87\% & 72.52\% & 73.14\% \\ 
         \hline
         SoNaR, \textit{windowed} & 86.36\% & 85.08\% & 86.26\%/ & \textbf{91.73\%}/ & 88.89\%/ \\
          & & & 86.53\% & 78.44\% & 82.25\% \\ 
         \hline
         SoNaR, \textit{windowed, no\_boundaries} & 88.36\% & 88.15\% & \textbf{91.05\%}/ & 89.24\%/ & 90.12\%/ \\
          & & & 84.59\% & \textbf{87.06\%} & 85.77\% \\
         \hline
         \multicolumn{6}{|c|}{Multitask Classification Model: Feedforward Context Encoder (2)} \\
         \hline
         SoNaR, \textit{windowed} & 88.16\% & 87.79\% & 90.37\%/ & 89.70\%/ & 90.02\%/ \\
          & & & 84.93\% & 85.88\% & 85.37\% \\ 
         \hline
         SoNaR, \textit{windowed, no\_boundaries} & 88.36\% & 88.14\% & 90.99\%/ & 89.31\%/ & 90.13\%/ \\
          & & & 84.66\% & 86.97\% & 85.77\% \\ 
         \hline
         \multicolumn{6}{|c|}{Multitask Classification Model: BiLSTM Context Encoder (3)} \\
         \hline
         SoNaR, \textit{windowed} & 88.63\% & 87.93\% & 89.58\% & 91.58\% & 90.55\% \\
          & & & \textbf{87.15\%} & 84.28\% & 85.66\% \\ 
         \hline
         SoNaR, \textit{windowed, no\_boundaries} & \textbf{88.85\%} & \textbf{88.51\%} & 90.95\% & 90.29\% & \textbf{90.60\%} \\
          & & & 85.83\% & 86.73\% & \textbf{86.25\%} \\ 
         \hline
    \end{tabular}
    \caption{Performance of the three multitask classification models for \textit{die}/\textit{dat} prediction}
    \label{tab:mlt_diedat}
\end{table}

\begin{table}[]
    \centering
    \begin{tabular}{|c|c|c|c|c|c|}
        \hline
         Dataset & Accuracy & Balanced & Precision & Recall & F1 \\
          & & accuracy & sc/rp/dp & sc/rp/dp & sc/rp/dp \\
         \hline
        \multicolumn{6}{|c|}{Multitask Classification Model: BiLSTM (1)} \\
         \hline
         SoNaR, \textit{full} & 70.72\% & 70.66\% & 71.99\%/ & 73.92\%/ & 72.88\%/ \\
          & & & 63.34\%/ & 68.29\%/ & 65.65\%/ \\
          & & & 75.30\% & 69.76\% & 72.38\% \\ 
         \hline
         SoNaR, \textit{windowed} & 83.15\% & 82.68\% & 84.35\%/ & 86.98\%/ & 85.61\%/ \\
          & & & 79.42\%/ & 76.92\%/ & 78.09\%/ \\
          & & & 84.53\% & 84.15\% & 84.31\% \\ 
         \hline
         SoNaR, \textit{windowed, no\_boundaries} & 85.69\% & 85.42\% & 88.78\%/ & 87.09\%/ & 87.90\%/ \\
          & & & 79.88\%/ & 82.49\%/ & 81.11\%/ \\ 
          & & & 87.24\% & 86.68\% & 86.93\% \\ 
         \hline
         \multicolumn{6}{|c|}{Multitask Classification Model: Feedforward Context Encoder (2)} \\
         \hline
         SoNaR, \textit{windowed} & 86.46\% & 86.14\% & 89.00\%/ & 87.80\%/ & 88.37\%/ \\
          & & & 80.24\%/ & 82.88\%/ & 81.49\%/ \\
          & & & 88.71\% & 87.75\% & 88.20\% \\ 
         \hline
         SoNaR, \textit{windowed, no\_boundaries} & 84.79\% & 84.76\% & 88.58\%/ & 86.23\%/ & 87.35\%/ \\
          & & & 77.04\%/ & 83.73\%/ & 80.19\%/ \\ 
          & & & 87.48\% & 84.31\% & 85.84\% \\ 
         \hline
         \multicolumn{6}{|c|}{Multitask Classification Model: BiLSTM Context Encoder (3)} \\
         \hline
         SoNaR, \textit{windowed} & \textbf{87.73\%} & \textbf{87.38\%} & \textbf{90.12\%} & \textbf{88.47\%} & \textbf{89.26\%} \\
          & & & \textbf{82.63\%} & 84.12\% & \textbf{83.31\%} \\
          & & & \textbf{89.27\%} & \textbf{89.55\%} & \textbf{89.39\%} \\ 
         \hline
         SoNaR, \textit{windowed, no\_boundaries} & 85.51\% & 85.48\% & 87.99\% & 86.98\% & 87.45\% \\
          & & & 78.90\% & \textbf{84.41\%} & 81.51\% \\ 
          & & & 88.31\% & 85.04\% & 86.61\% \\ 
         \hline
    \end{tabular}
    \caption{Performance results of three multitask classification tasks for POS prediction: \textit{subordinating conjunction}(sc), \textit{relative pronoun} (rp) and \textit{demonstrative pronoun} (dp)}
    \label{tab:mlt_functions}
\end{table}

\section{Discussion}

In Section \ref{binary_classification_model}, a first classification model is computed to predict \textit{die} and \textit{dat} labels. The binary classification model (Model 1) consists of an embedding layer, a bidirectional LSTM, a maxpooling layer and a linear layer. The softmax is taken over the output of the last layer and provides a probability distribution over \textit{die} and \textit{dat} prediction labels. The sentences receive the prediction label with the highest probability. It is trained, validated and tested four times using four different database settings. From an analysis of the performance metric results, several conclusions can be drawn. Firstly, in all cases, the model appears to predict the \textit{dat} label more precisely than the \textit{die} label. This may be caused by the higher number of \textit{dat} than \textit{die} instances in training, validation and test datasets extracted from the Europarl and SoNaR corpus. Secondly, when the dataset is more balanced, as in the SoNaR corpus, the difference in performance between \textit{die} and \textit{dat} labels decreases as expected. Thirdly, \textit{die/dat} prediction performance increases when the window over the sentences is not limited to sentence boundaries (SoNaR \textit{windowed, no\_boundaries}). A probable reason for that higher performance is that the model is able to detect antecedents in the preceding or following sentence, while it is not able to do so when it is trained and tested on boundary-constraint windowed sentences (SoNaR \textit{windowed}). Lastly, it appears that performance of the model drops significantly when the binary classification model is trained and tested on full sentences (Europarl \textit{full}). In conclusion, the binary classification model performs best when it is trained on the larger, more evenly balanced SoNaR corpus that consists of windowed sentences that are not limited to sentence boundaries. A clear performance overview of the best performing binary classification and multitask classification models for \textit{die/dat} prediction can be found in Table \ref{tab:comparison1}.

\begin{table}[]
    \centering
    \begin{tabular}{|c|c|c|c|c|}
        \hline
        \multicolumn{5}{|c|}{Best performing models: die/dat prediction} \\
        \hline
        \textit{die/dat} & Model 1 & Model 2 & Model 3 & Model 4 \\
        \hline
        \hline
        Accuracy & 84.56\% & 88.36\% & 88.16\% & \textbf{88.63\%} \\
        \hline
        Balanced Accuracy & 84.18\% & \textbf{88.15\%} & 87.14\% & 87.93\% \\
        \hline
        \hline
        \textit{dat} (0) &  & &  &  \\
        \hline
        Precision & 87.71\% & \textbf{91.05\%} & 90.37\% & 89.58\% \\
        \hline
        Recall & 86.16\% & 89.24\% & 89.70\% & \textbf{91.58\%} \\
        \hline
        F1-score & 86.85\% & 90.12\% & 90.02\% & \textbf{90.55\%} \\
        \hline
        \hline
        \textit{die} (1) & & & & \\
        \hline
        Precision & 80.13\% & 84.59\% & 84.66\% & \textbf{87.15\%} \\
        \hline
        Recall & 82.20\% & \textbf{87.06\%} & 86.97\% & 84.28\% \\
        \hline
        F1-score & 80.99\% & \textbf{85.77\%} & \textbf{85.77\%} & 85.66\% \\
        \hline
    \end{tabular}
    \caption{Comparison of \textit{die/dat} prediction performance between best performing binary classification model (model 1, SoNaR \textit{windowed, no\_boundaries}), multitask classification model (model 2, SoNaR \textit{windowed, no\_boundaries}), multitask classification model with feedforward context encoder (model 3, SoNaR \textit{windowed}) and multitask classification model with bidirectional LSTM context encoder (model 4, SoNaR \textit{windowed})}
    \label{tab:comparison1}
\end{table}

In Section \ref{multitask_classification_model}, three multitask classification models are constructed to jointly execute two prediction tasks: \textit{die/dat} prediction and POS prediction. The BiLSTM multitask classification model (Model 2) consists of an embedding layer, two consecutive BiLSTMs and a maxpooling layer. The output of the maxpooling layer is used as input to two separate linear layers followed by a softmax layer. The two softmax layers yield a probability distribution for \textit{die/dat} and POS labels. The model trained and tested on windowed SoNaR sentences that exceed sentence boundaries performs better than the model on boundary-constraint windowed sentences and full sentences. The best performing BiLSTM multitask classification model (Model 2) outperforms the best binary classification model (Model 1) on every evaluation metric for \textit{die/dat} prediction. This could arguably be due to the increased batch size, the doubled embedding dimension, the extra bidirectional LSTM layer, the influence of the second prediction task and/or the split in sentence and context encoder. Firstly, we test the influence of the increased batch size. For this, we retrain the multitask classification model and feed the data in batches of 128 (used for binary classifier training) instead of 512 samples. Table \ref{tab:batchsize} consistently shows that there is little consistent difference in performance when batch size is 512 or 128. Therefore, it can be suggested that an increased batch size has no directly positive influence on model performance. Secondly, we retrain the multitask classification model and let the embedding layer transform the input data to 100-dimensional word embeddings instead of 200-dimensional word embeddings. From the results displayed in Table \ref{tab:batchsize}, it appears that an increase in word embedding dimension does indeed cause a slight increase in model performance. Thirdly, the multitask model contains two BiLSTM layers opposed to the binary model that has only one layer. Table \ref{tab:layers} shows the influence of the number of layers on the performance of the binary classification model. When the binary classification model is retrained with an additional BiLSTM layer, all the evaluation metrics rise with approximately 2\%. However, when the binary classification model has three BiLSTM layers, model performance drops significantly. It appears that the doubled number of layers is indeed one of the reasons why the multitask classification models perform better than the binary classification model. However, not every rise in number of layers necessarily influences a model's performance in a positive manner. Concerning the influence of the POS prediction task on \textit{die/dat} prediction performance, a comparison between a two-layer BiLSTM binary classification model (Model 1) and the two-layer BiLSTM multitask classification model (Model 2) is made and displayed in Table \ref{tab:linguisticknowledge}. It seems that the integration of POS knowledge positively influences \textit{die/dat} prediction performance, as all evaluation metrics have increased. When examining the influence of a context encoder on \textit{die/dat} prediction performance, the evaluation metrics of Model 2, 3 and 4 are compared. The results of the three models are fairly similar which leads to the conclusion that the addition of a context encoder has little to no further influence on \textit{die/dat} prediction performance. Moreover, the encoder architecture does not cause a considerable difference in \textit{die/dat} prediction performance between the model with a feedforward context encoder (Model 3) and the model with a BiLSTM context encoder (Model 4). It can thus be suggested that a model does not necessarily profit from a different architecture and that an extra focus on immediate context is not additionally advantageous for the \textit{die/dat} prediction task.

\begin{table}[]
    \centering
    \begin{tabular}{|c|c|c|c|c|c|}
        \hline
        \multicolumn{6}{|c|}{Batch Size/Embedding Dimension} \\
        \hline
         Batch size/ & Accuracy & Balanced & Precision & Recall & F1 \\
         Embedding & & accuracy & dat/die & dat/die & dat/die \\
         \hline
         \hline
         512/200 & 88.36\% & 88.15\% & 91.05\% & 89.24\% & 90.12\% \\
          & & & 84.59\% & 87.06\% & 85.77\% \\ 
         \hline
         128/200 & 87.46\% & 88.73\% & 89.43\% & 91.45\% & 90.37\% \\
          & & & 86.94\% & 84.02\% & 85.33\% \\ 
         \hline
         512/100 & 86.94\% & 87.77\% & 88.54\% & 91.29\% & 89.88\% \\
          & & & 86.54\% & 82.58\% & 84.48\% \\ 
         \hline
    \end{tabular}
    \caption{The influence of batch size and embedding dimension on performance of the SoNaR-based, sentence-exceeding windowed trained multitask classification model (Model 2, SoNaR \textit{windowed, no\_boundaries})}
    \label{tab:batchsize}
\end{table}

\begin{table}[]
    \centering
    \begin{tabular}{|c|c|c|c|c|c|}
        \hline
        \multicolumn{6}{|c|}{Number of layers} \\
        \hline
         Layers & Accuracy & Balanced & Precision & Recall & F1 \\
          & & accuracy & dat/die & dat/die & dat/die \\
         \hline
         \hline
         1 & 84.56\% & 84.18\% & 87.71\% & 86.16\% & 86.85\% \\
          & & & 80.13\% & 82.20\% & 80.99\% \\ 
         \hline
         2 & 87.21\% & 86.83\% & 89.62\% & 88.82\% & 89.15\% \\
          & & & 83.76\% & 84.84\% & 84.16\% \\ 
         \hline
         3 & 75.75\% & 76.89\% & 80.01\% & 81.54\% & 80.74\% \\
          & & & 72.02\% & 69.97\% & 70.93\% \\ 
         \hline
    \end{tabular}
    \caption{The influence of number of layers on performance of the SoNaR-based, sentence-exceeding windowed trained binary classification model (Model 1, SoNaR \textit{windowed, no\_boundaries})}
    \label{tab:layers}
\end{table}

\begin{table}[]
    \centering
    \begin{tabular}{|c|c|c|c|c|c|}
        \hline
        \multicolumn{6}{|c|}{Integrated POS knowledge} \\
        \hline
         Linguistic & Accuracy & Balanced & Precision & Recall & F1 \\
         classes & & accuracy & dat/die & dat/die & dat/die \\
         \hline
         \hline
         Yes & 88.36\% & 88.15\% & 91.05\% & 89.24\% & 90.12\% \\
          & & & 84.59\% & 87.06\% & 85.77\% \\ 
         \hline
         No & 87.21\% & 86.83\% & 89.62\% & 88.82\% & 89.15\% \\
          & & & 83.76\% & 84.84\% & 84.16\% \\ 
         \hline
    \end{tabular}
    \caption{The influence of integrated POS knowledge on \textit{die/dat} prediction performance. Comparison between Model 1 with an extra BiLSTM layer (\textit{No}) and Model 2 (\textit{Yes}), both trained and tested using SoNaR \textit{windowed, no\_boundaries} dataset}
    \label{tab:linguisticknowledge}
\end{table}

\begin{table}[]
    \centering
    \begin{tabular}{|c|c|c|c|}
        \hline
        \multicolumn{4}{|c|}{Best performing models: POS prediction} \\
        \hline
        \textit{linguistic classes} & Model 2 & Model 3 & Model 4 \\
        \hline
        \hline
        Accuracy & 85.69\% & 86.46\% & \textbf{87.73\%} \\
        \hline
        Balanced Accuracy & 85.42\% & 86.14\% & \textbf{87.38\%} \\
        \hline
        \hline
        \textit{subordinating conjunction} (0) &  &  & \\
        \hline
        Precision & 88.78\% & 89.00\% & \textbf{90.12\%} \\
        \hline
        Recall & 87.09\% & 87.80\% & \textbf{88.47\%} \\
        \hline
        F1-score & 87.90\% & 88.37\% & \textbf{89.26\%} \\
        \hline
        \hline
        \textit{relative pronoun} (1) &  & &  \\
        \hline
        Precision & 79.88\% & 80.24\% & \textbf{82.63\%} \\
        \hline
        Recall & 82.49\% & 82.88\% & \textbf{84.12\%} \\
        \hline
        F1-score & 81.11\% & 81.49\% & \textbf{83.31\%} \\
        \hline
        \hline
        \textit{demonstrative pronoun} (2) & & & \\
        \hline
        Precision & 87.24\% & 88.71\% & \textbf{89.27\%} \\
        \hline
        Recall & 86.68\% & 87.75\% & \textbf{89.55\%} \\
        \hline
        F1-score & 86.93\% & 88.20\% & \textbf{89.39\%} \\
        \hline
    \end{tabular}
    \caption{Comparison of POS prediction performance between best performing multitask classification model (model 2, SoNaR \textit{windowed, no\_boundaries}), multitask classification model with feedforward context encoder (model 3, SoNaR \textit{windowed}) and multitask classification model with bidirectional LSTM context encoder (model 4, SoNaR \textit{windowed})}
    \label{tab:comparison2}
\end{table}

Contrary to the little to no impact it has on \textit{die/dat} prediction performance, the context encoder - especially the BiLSTM context encoder - does have a direct positive impact on POS prediction performance. The difference in POS prediction performance between the three multitask prediction models can be found in Table \ref{tab:comparison2}. The model with the BiLSTM context encoder (Model 4) outperforms the other two multitask classification models on every evaluation metric. Considering its highest POS prediction performance and high \textit{die/dat} prediction performance, it can be concluded that the multitask prediction model with BiLSTM context encoder (Model 4) is the overall best model.

\section{Conclusion and Future Work}

Deciding which pronoun to use in various contexts can be a complicated task. The correct use of \textit{die} and \textit{dat} as Dutch pronouns entails knowing the linguistic class of the antecedent and - if the antecedent is a noun - its grammatical gender and number. We experimented with neural network models to examine whether \textit{die} and \textit{dat} instances in sentences can be computationally predicted and, if necessary, corrected. Our binary classification model reaches a promising 84.56 \% accuracy. In addition, we extended the model to a multitask model which apart from the \textit{die} and \textit{dat} prediction also predicts their POS (\textit{demonstrative pronoun}, \textit{relative pronoun} and \textit{subordinating conjunction}). By increasing the word embedding dimension, doubling the number of bidirectional LSTM layers and integrating POS knowledge in the model, the multitask classification models raise \textit{die/dat} prediction performance by approximately 4 \%. Concerning POS prediction performance, the multitask classification model consisting of a sentence and context encoder performs best on all evaluation metrics and reaches 87.78 \% accuracy. 

There are ample opportunities to further analyze, enhance and/or extend the \textit{die/dat} prediction model. A qualitative study of the learned model weights, for example, could provide more insight in the prediction mechanism of the models. We already obtain excellent results with a simple neural architecture comprising relatively few parameters. We believe that more complex architectures such as a transformer architecture \cite{vaswani2017attention} with multihead attention will improve results. It might also be interesting to look at the possibility of integrating a language model such as BERT \cite{devlin2018bert} in the classification model (e.g., as pretrained embeddings). Moreover, the binary classification task could be extended to a multiclass classification task to predict not only \textit{die} and \textit{dat} labels, but also respectively equivalent \textit{deze} and \textit{dit} labels. The difference between \textit{die/dat} and \textit{deze/dat}, however, entails a difference in temporal and spatial information: while \textit{die/dat} indicates a physically distant or earlier mentioned antecedent, \textit{deze/dit} implies that the antecedent is physically near or later mentioned in the text. Moreover, \textit{die/dat} and \textit{deze/dit} are preferably used for anaphoric and cataphoric reference, respectively. The difference in reference (examples 1 and 2) and spatial understanding (example 4) between \textit{dat/dit} and \textit{die/deze} is demonstrated below.

\begin{enumerate}
    \item Je bent gek. \textit{Dat} heb ik je al gezegd. ("You are crazy. I have told you \textit{that} already.") \cite{vrttaal}
    \item Ik heb je \textit{dit} al gezegd: je bent gek. ("I have to tell you \textit{this}: you are crazy.")
    \item Ik heb je al gezegd \textit{dat} je gek bent. ("I have told you already \textit{that} you are crazy.")
    \item Lees eerst \textit{deze} boeken, dan \textit{die} andere. ("First, read \textit{these} books, than \textit{those} other.") \cite{taaltelefoon}
\end{enumerate}

\textit{Dat} in example 1 indicates an anaphoric reference to the previous sentence. The same message is conveyed in example 2, but the sentence is referred to cataphorically using \textit{dit}. Example 3 is very similar to example 2 in terms of sequence in which the information is provided. However, \textit{dat} and \textit{dit} differ in POS: \textit{dit} is an independent demonstrative pronoun and functions as direct object (example 2), whereas \textit{dat} is a subordinating conjunction and the entire subordinate clause "\textit{dat je gek bent}" functions as direct object (example 3). In addition, the word order differs in both examples. Finally, \textit{deze} (example 4) indicates that its antecedent is spatially close to the speaker, whereas \textit{die} is spatially distant. In order to learn the difference between \textit{dat/dit} and \textit{die/deze}, the model may need to focus more on the antecedent's position with respect to the pronoun, POS, word order and other tokens in the sentences such as colons, and it will need to infer the spatial (and temporal) relation between the speaker and the antecedent. 


\bibliographystyle{clin} 
\bibliography{bibliography}  

\end{document}